\pdfoutput=1
\documentclass[letterpaper, 10 pt, conference]{ieeeconf}  

\IEEEoverridecommandlockouts                              

\usepackage{mathptmx} 
\usepackage{amsmath} 
\usepackage{amssymb}  
\usepackage{graphicx}
\usepackage{subcaption}
\usepackage{url}
\usepackage{svg}
\usepackage{textcomp}
\usepackage{xcolor}
\usepackage{dblfloatfix}

\title{\LARGE \bf
Hierarchical Mask2Former: Panoptic Segmentation of Crops, Weeds and Leaves
}

\author{Madeleine Darbyshire$^{1}$ and Elizabeth Sklar$^{2}$ and Simon Parsons$^{1}$
\thanks{$^{1}$MD and SP are in the School of Computer Science, University of Lincoln, UK.
        {\tt 25696989@students.lincoln.ac.uk, sparsons@lincoln.ac.uk}}
\thanks{$^{2}$ ES is with the Lincoln Institute of Agri-food Technology, University of Lincoln, UK.
        {\tt esklar@lincoln.ac.uk}}
}

\begin{document}

\maketitle
\thispagestyle{empty}
\pagestyle{empty}

\begin{abstract}
Advancements in machine vision that enable detailed inferences to be made from images have the potential to transform many sectors including agriculture. \emph{Precision agriculture}, where data analysis enables interventions to be precisely targeted, has many possible applications. Precision spraying, for example, can limit the application of herbicide only to weeds, or limit the application of fertiliser only to undernourished crops, instead of spraying the entire field. The approach promises to maximise yields, whilst minimising resource use and harms to the surrounding environment. To this end, we propose a hierarchical panoptic segmentation method to simultaneously identify indicators of plant growth and locate weeds within an image. We adapt Mask2Former, a state-of-the-art architecture for panoptic segmentation, to predict crop, weed and leaf masks. We achieve a \(PQ^\dagger\) of 75.99. Additionally, we explore approaches to make the architecture more compact and therefore more suitable for time and compute constrained applications. With our more compact architecture, inference is up to 60\% faster and the reduction in \(PQ^\dagger\) is less than 1\%. 
\end{abstract}


\section{INTRODUCTION}
The continued growth of the global population has put farmers are under pressure to produce more food to meet the increasing demand. However, concerns about the environmental impact of agriculture are placing a simultaneous pressure on farmers to mitigate the environmental harms of their operations. All the while, climate changes is making growing conditions more unpredictable leading to new challenges in providing a reliable supply of food. 

\emph{Precision agriculture} aims to leverage data and machine learning to help farmers make informed decisions, and target interventions precisely. For example, herbicide usage can be reduced by first detecting and then only targeting weeds rather than spraying the entire field with herbicide. Furthermore, crop monitoring can indicate where fertiliser should be targeted for healthy plant growth. Various phenotypic traits can be used as indicators of crop growth, but in this paper we use leaf count.

The paper aims to combine crop and weed segmentation as well as leaf segmentation in a single hierarchical panoptic segmentation architecture. Our approach adapts the latest state-of-the-art panoptic segmentation architecture to the task and improves upon existing baselines. Moreover, we propose ways to make the model more compact, with faster inference times, whilst retaining accuracy \footnote{Our code is published at: \url{https://github.com/madeleinedarbyshire/HierarchicalMask2Former}}.

\begin{figure}[t]
\centering
\begin{subfigure}{0.99\columnwidth}
    \centering
    \includegraphics[width=0.45\linewidth]{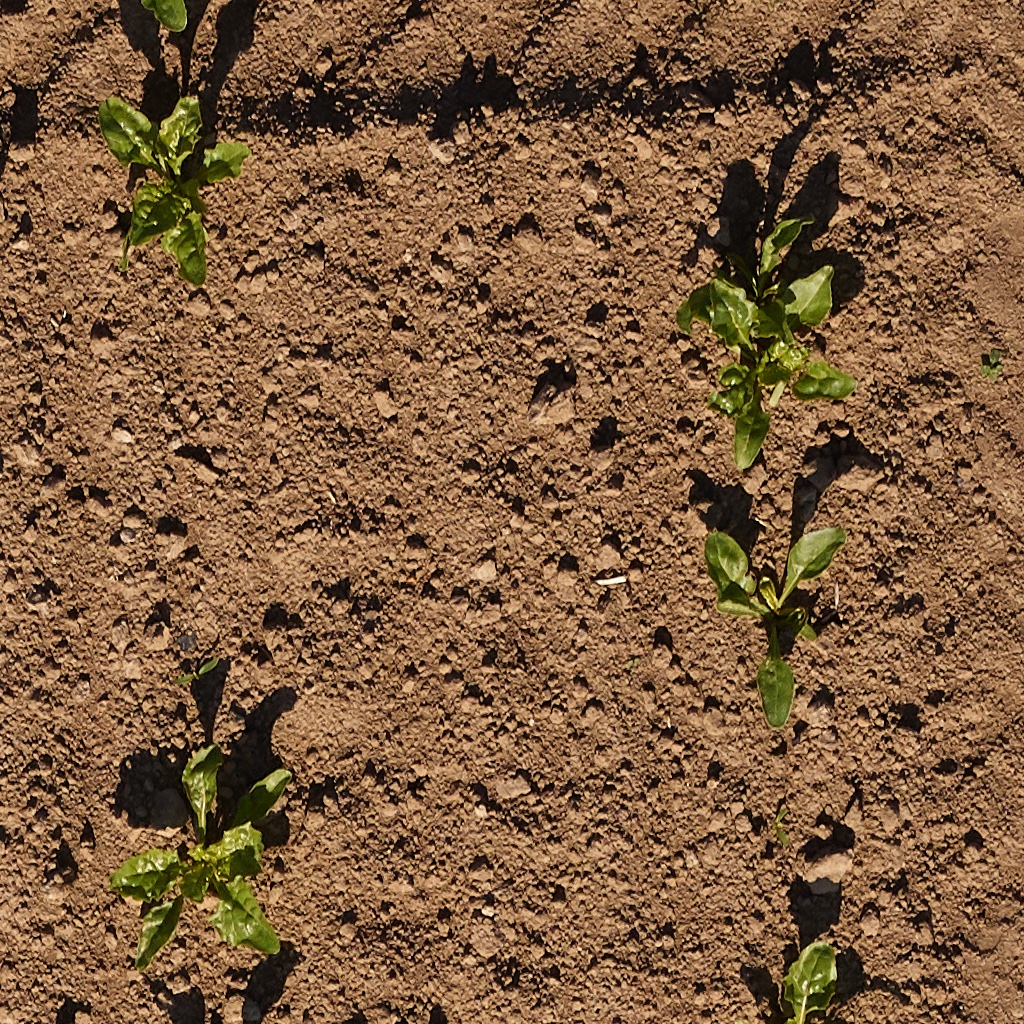}
    \includegraphics[width=0.45\linewidth]{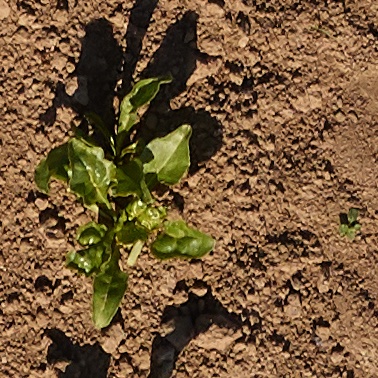}
    \caption{Original}
\end{subfigure}\vspace{0.3cm}
\begin{subfigure}{0.99\columnwidth}
    \centering
    \includegraphics[width=0.45\linewidth]{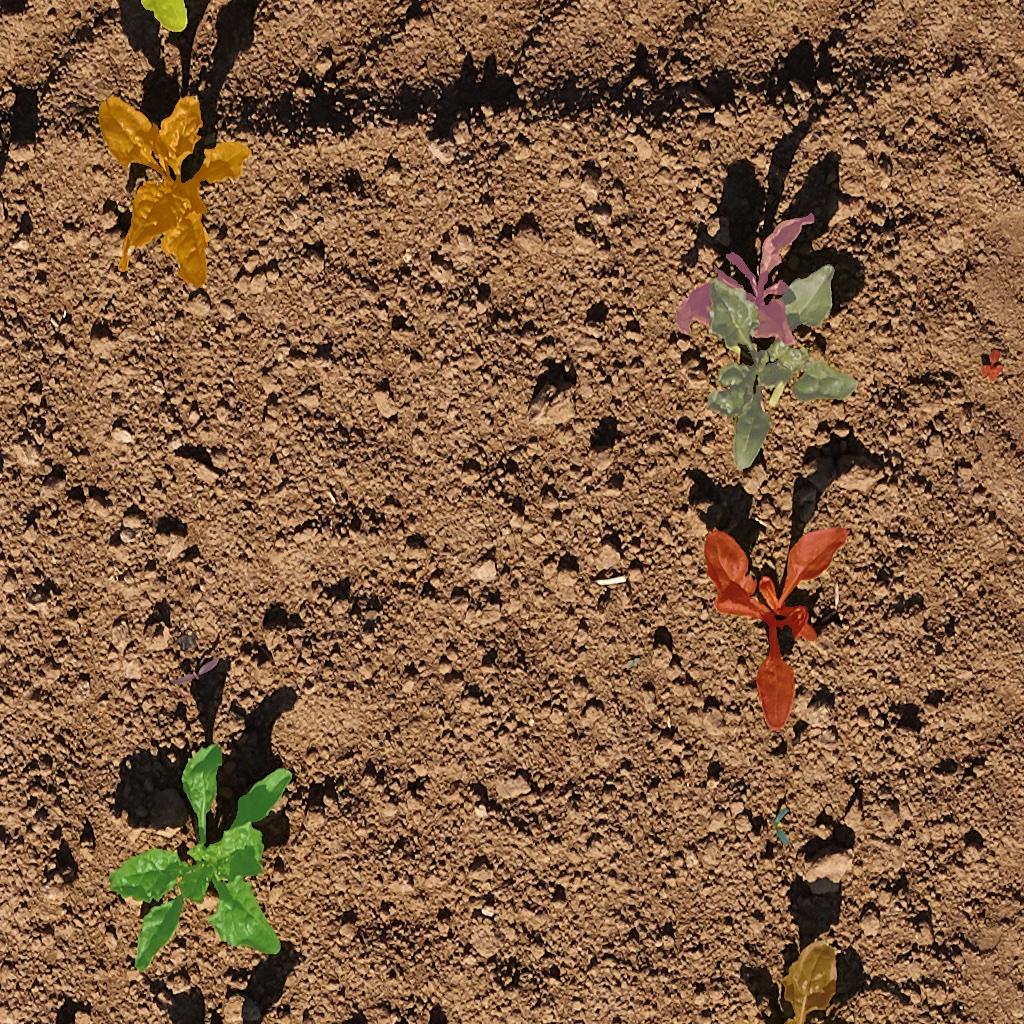}
    \includegraphics[width=0.45\linewidth]{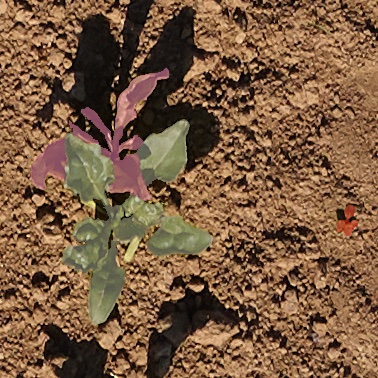}
    \caption{Plant Instances}
\end{subfigure}\vspace{0.3cm}
\begin{subfigure}{0.99\columnwidth}
    \centering
    \includegraphics[width=0.45\linewidth]{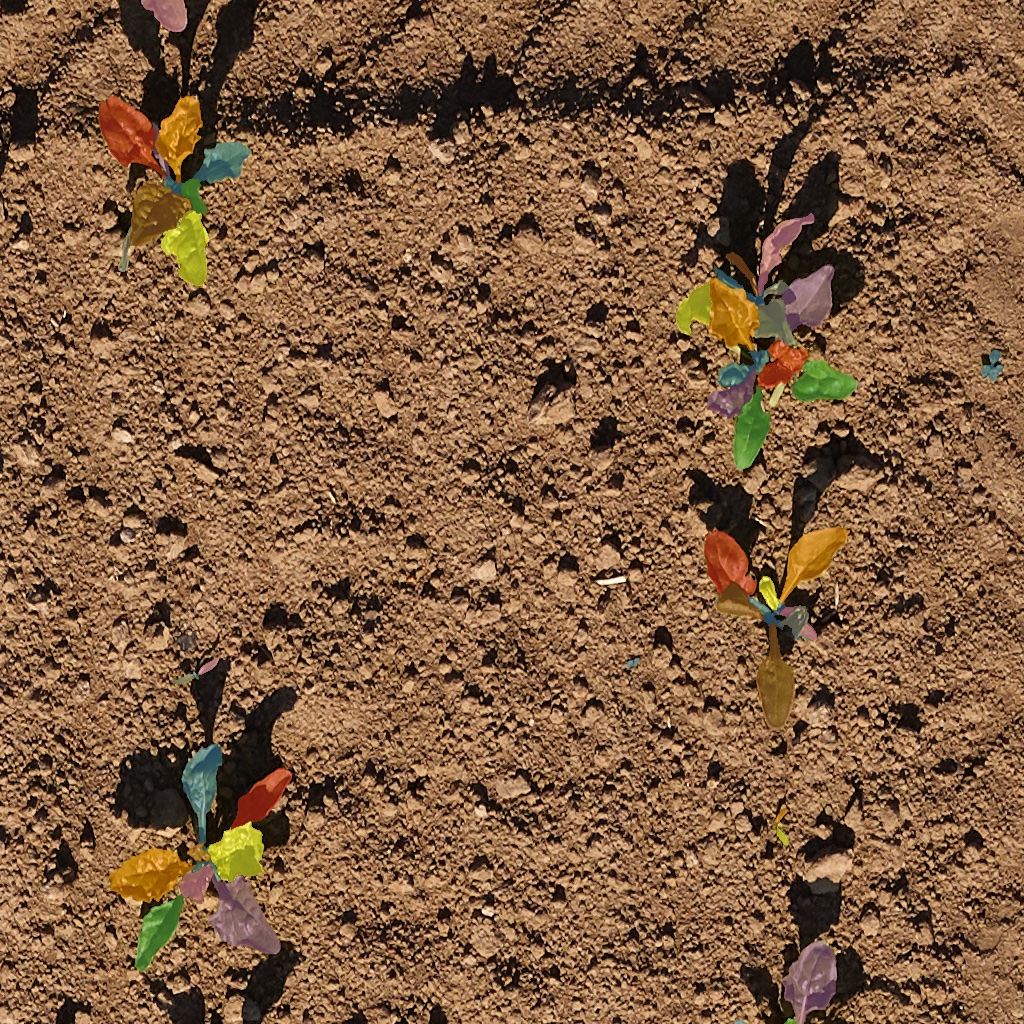}
    \includegraphics[width=0.45\linewidth]{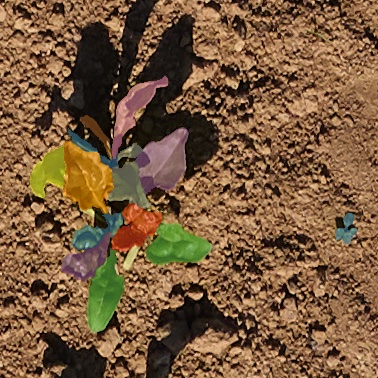}
    \caption{Leaf Instances}
\end{subfigure}
\caption{Example images from the PhenoBench dataset. On the left are the full size images and the right shows a close-up of the same image.}
\label{fig:examples}
\end{figure}

\section{RELATED WORK}
In recent years, most approaches to crop and weed segmentation have utilised convolutional neural networks (CNNs) in one form or another. Early examples used SegNet \cite{badrinarayanan2017segnet} to classify each pixel in the image as either crop, weed or background \cite{di2017automatic, sa2018weednet}.  SegNet employs an encoder-decoder structure, where the encoder extracts hierarchical features from input images, and the decoder produces pixel-wise segmentation masks.  This encoder-decoder segmentation architecture was further improved in DeepLabV3+ \cite{chen2017rethinking} which added atrous convolutions to capture larger spatial context. Moreover, U-Net \cite{ronneberger2015u}, another similar segmentation approach, introduces skip connections that allow it to capture both high-level and low-level features. In \cite{zou2021modified} and \cite{genze2022deep}, U-Net outperforms DeepLabV3+ in crop and weed segmentation.

Beyond classifying the pixels, instance segmentation has been used to distinguish between individual crop and weed plants. Mask R-CNN \cite{osorio2020deep}, a widely used instance segmentation technique, combines pixel-level semantic segmentation with object bounding box predictions to segment instances. This approach can be applied to many panoptic segmentation problems as well. Panoptic-DeepLab \cite{cheng2020panoptic} builds on an adapted version of DeepLabV3+, adding instance segmentation segmentation heads, to make it suitable for instance segmentation and panoptic segmentation. Mask2Former \cite{cheng2021mask2former}, and its predecessor MaskFormer \cite{cheng2021maskformer}, utilise an approach to instance and panoptic segmentation that differs from these per-pixel approaches. Instead, images are partitioned into a number of regions, represented with binary masks, then each of these is assigned a class.

Adjacent to the problem of identifying individual plant instances is the identification of individual leaf instances within each plant instance. A catalyst for research in this area was the CVPPP Leaf Segmentation Challenge \cite{scharr2014annotated}. The goal of which was to extract plant traits from images of single plants in laboratory conditions. One successful approach was \cite{aich2017leaf} where SegNet was used to create binary masks of the plants and a regression network was used to count the leaves. More recently, \cite{weyler2022joint} and \cite{weyler2022field} demonstrated each plant and its leaves could be identified from images, containing multiple plants, taken under real field conditions. Subsequently, the tasks of crop and weed segmentation, and leaf instance segmentation were combined in \cite{roggiolani2023hierarchical}. In \cite{roggiolani2023hierarchical}, ERFNet \cite{romera2017erfnet} is adapted and a second decoder is added so that there is one decoder for generating plant masks and another for generating leaf masks.

In this paper we explore whether the segmentation of crops, weeds and leaves can be improved by adapting state-of-the-art panoptic segmentation architecture, Mask2Former. We explore different configurations to understand how best to adapt it to the task of hierarchical segmentation. Additionally, we consider how to make the architecture more compact and therefore more suitable for compute and time-constrained applications.

\section{METHODOLOGY}

\subsection{Mask2Former}
\begin{figure}[!ht]
\centering
\includegraphics[width=0.49\textwidth]{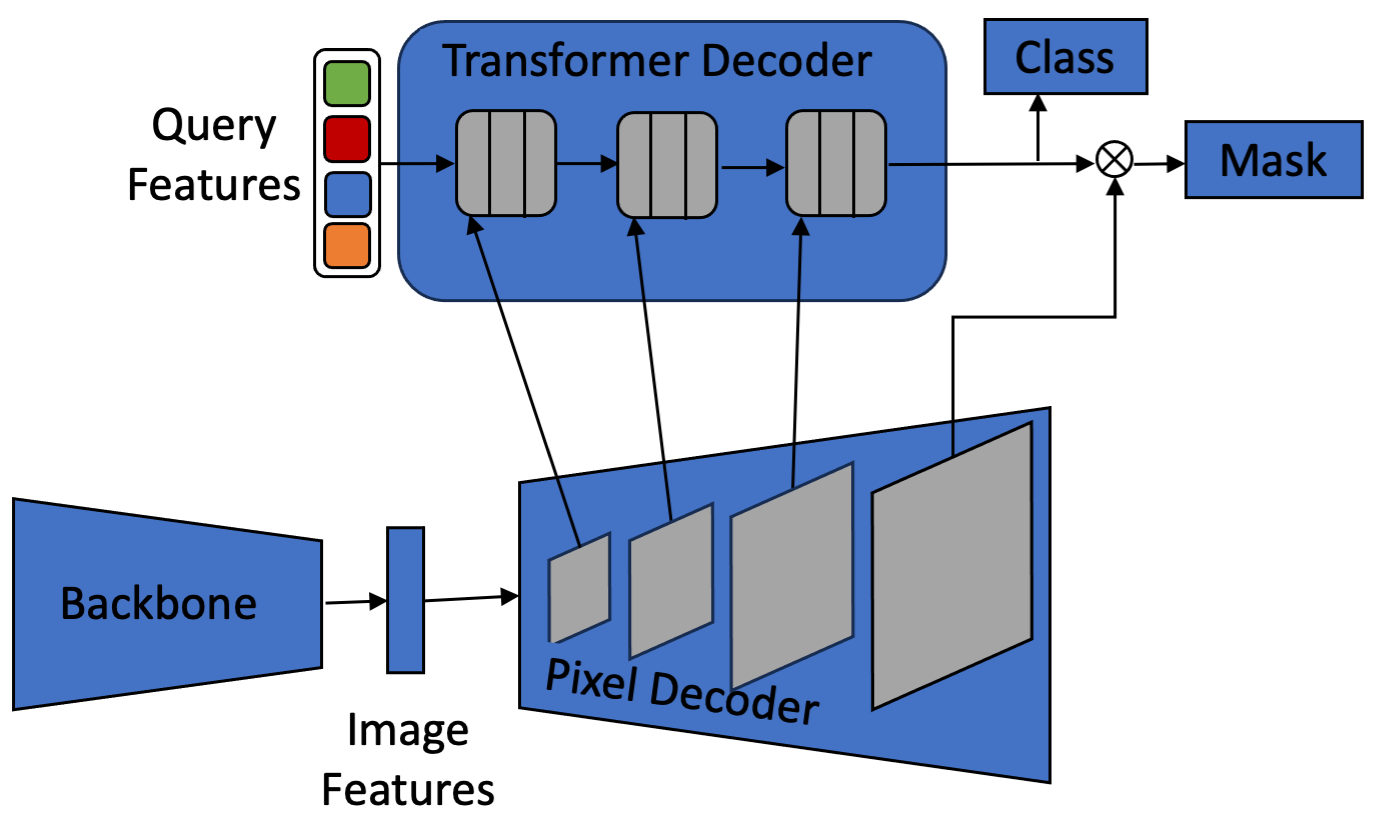}
\caption{Original Mask2Former architecture. Each grey block in the transformer decoder contains 3 transformer layers that correspond to each scale in the pixel decoder's feature pyramid.}
\label{fig:m2f}
\end{figure}

Our approach adapts the Mask2Former architecture. The original Mask2Former architecture is shown in Figure \ref{fig:m2f} and consists of the following:

\textbf{Backbone.} The backbone extracts low-level image features $ F^{{C_F} \times \frac{H}{S} \times \frac{W}{S}} \in \mathbb{R} $ from an input image of size $H\times W$, where $C_F$ is the number of channels and $S$ is the stride.

\textbf{Pixel decoder.} The pixel decoder gradually upsamples the low-level features to produce a feature pyramid with layers that are of resolution 1/32, 1/16 and 1/8. 
At each stage in the upsampling process, a per-pixel embedding is created $\epsilon_{pixel} \in \mathbb{R}^{C_{\epsilon} \times H \times W}$, where $C_{\epsilon}$ is the embedding dimension. In this implementation, the advanced multi-scale deformable attention Transformer, \emph{MSDeformAttn} \cite{zhu2020deformable} is used as the pixel decoder. 

\textbf{Transformer decoder.} The transformer decoder consists of 3 transformer layers for each layer of the feature pyramid. Therefore, given there are 3 layers in the feature pyramid, there are 9 transformer decoder layers. Each transformer decoder layer consists of a self-attention layer, a cross-attention layer and a feed-forward network. Query features are associating with the positional embeddings produced by the pixel decoder. These query features are learnable and thus updated by each layer of the network. The transformer outputs $N$ per-segment embeddings, $Q^{C_Q \times N} \in \mathbb{R}$, where $N$ is the number of queries and $C_Q$ is the dimension that encodes global information about the segment. 

\textbf{Segmentation Module}
The segmentation module transforms the output of the transformer $Q$ into masks and class predictions. To acquire class probability predictions $\{p_i \in \Delta^{K}\}^N_{i=1}$, a linear classifier and softmax activation are applied to the output. There is an additional no-object class which applies where the embedding does not correspond to any region. To generate the masks, a multi-layer perceptron converts the per-segment embeddings from the transformer into $N$ mask embeddings $\epsilon_{mask} \in \mathbb{R}^{C_\epsilon \times N}$. Lastly, binary masks $m_i \in [0, 1]^{H \times W}$ are formed via the dot product of the mask embeddings, $\epsilon_{mask}$, and per-pixel embeddings, $\epsilon_{pixel}$, followed by a sigmoid activation $m_i[h,w] = sigmoid(\epsilon_{mask}[:, i]^T \cdot \epsilon_{pixel}[:,h,w])$.

\subsection{Separate Segmentation Modules for Plants and Leaves}
\label{method:separate_seg}
One lightweight approach to adapt Mask2Former is have separate segmentation modules for plants and leaves. Each module takes the output of the transformer decoder, $Q$, and forms two sets of class predictions $\{p^{plant}_i \in \Delta^{K_{plant}}\}^N_{i=1}$ and $\{p^{leaf}_i \in \Delta^{K_{leaf}}\}^N_{i=1}$ for plants and leaves, respectively. Additionally, separate mask embeddings $\epsilon^{plant}_{mask} \in \mathbb{R}^{C_\epsilon X N}$ and $\epsilon^{leaf}_{mask} \in \mathbb{R}^{C_\epsilon X N}$ are generated via separate multi-layer perceptrons. Each of these, combined with the pixel embedding $\epsilon_{pixel}$ using the dot product would produce the two sets of binary mask predictions: $m^{plant}_i \in [0, 1]^{H \times W}$ and $m^{leaf}_i \in [0, 1]^{H \times W}$ using the same approach described above. 

For plant masks:
\begin{equation}
    m^{plant}_i[h,w] = sigmoid(\epsilon^{plant}_{mask}[:, i]^T \cdot \epsilon_{pixel}[:,h,w])
\end{equation}

For leaf masks:
\begin{equation}
    m^{leaf}_i[h,w] = sigmoid(\epsilon^{leaf}_{mask}[:, i]^T \cdot \epsilon_{pixel}[:,h,w])
\end{equation}

\subsection{Separate Transformer Decoders for Plants and Leaves}
Another approach, is to use separate transformer decoders: one for the plants and one for the leaves. By having separate plant and leaf transformer decoders, we thought the network might learn better how to segment at the plant level versus at the leaf level. Each transformer decoder takes its own set of learnable query features and each produces N per-segment embeddings $Q_{plant}$ and $Q_{leaf}$, respectively. The output from the each transformer decoder is then converted to class predictions and masks via the separate segmentation modules, as defined in Section \ref{method:separate_seg}.

\subsection{Compact Transformer Decoder}
\begin{figure}[!ht]
\centering
\includegraphics[width=0.5\textwidth]{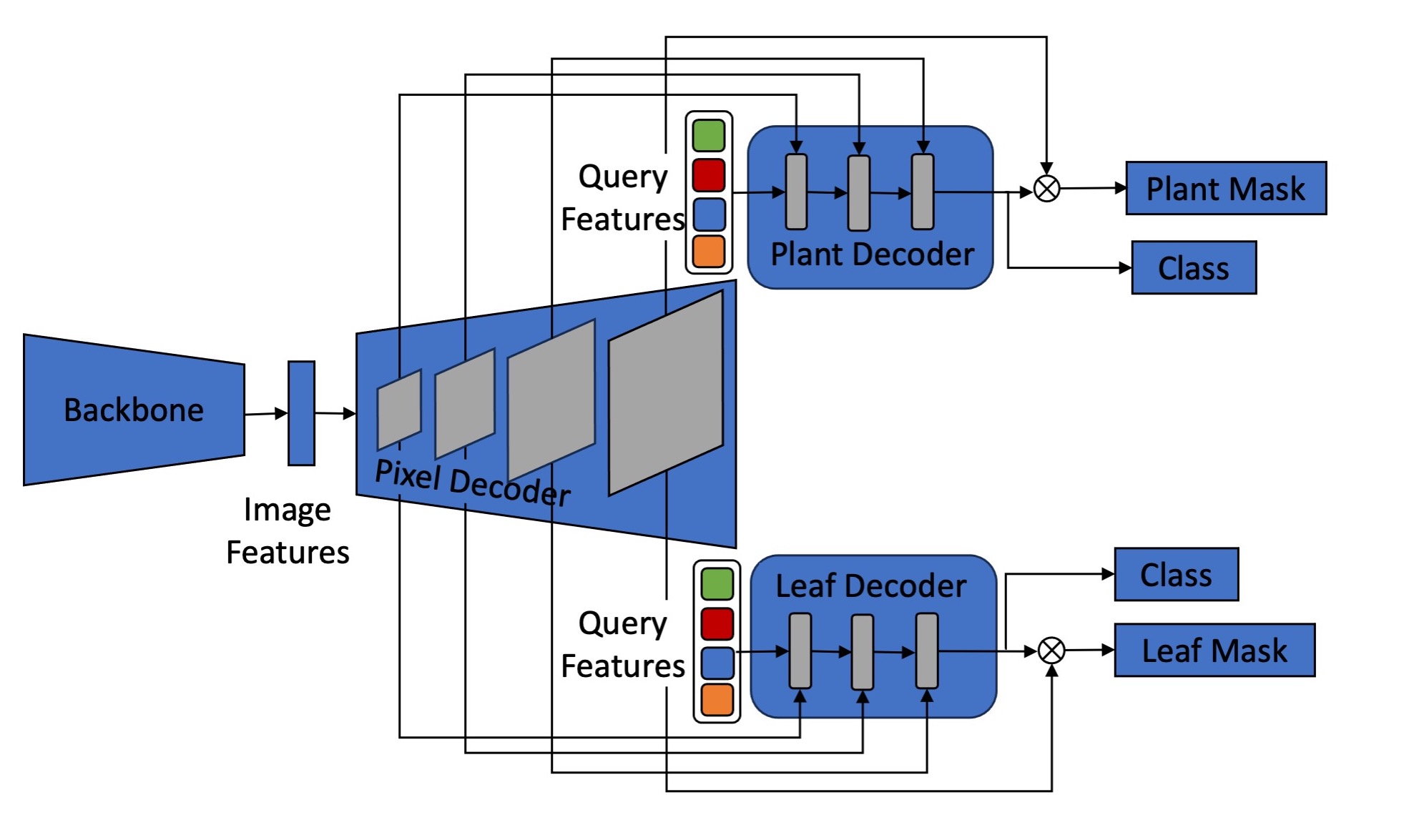}
\caption{Our adapted Mask2Former architecture with separate transformer decoders for plants and leaves, and one layer for each feature level instead of the three used in the original Mask2Former architecture.}
\label{fig:lobster}
\end{figure}

In order to make the architecture of Mask2Former more suitable for real-time applications, we trial a more compact transformer decoder architecture. Instead of three layers for each feature level, we reduce this to just one layer per feature level. In particular, we wanted to reduce the computational overhead of adding a second transformer. Figure \ref{fig:lobster} demonstrates this more compact transformer decoder in the two-transformer decoder design.

\subsection{Loss Function}
As in Mask2Former, the mask loss is calculated using the sum of the binary cross entropy loss and dice loss:

\begin{equation}
    L_{\text{mask}} = L_{\text{ce}} + L_{\text{dice}}
\end{equation}

The mask loss and class loss is calculated for both the plants and leaves,  respectively. The total loss is calculates as a weighted sum of the classification and mask loss of the plants and leaves:

\begin{equation}
    L = \lambda_{\text{cls}} L^{p}_{\text{cls}} + \lambda_{\text{mask}} L^{p}_{\text{mask}} + \lambda_{\text{cls}} L^{l}_{\text{cls}} + \lambda_{\text{mask}} L^{l}_{\text{mask}}
\end{equation}

where $L^p$ and $L^l$ are the losses for the plants and leaves respectively. The weights for each of the losses were set to $\lambda_{\text{mask}}=2.5$ and $\lambda_{\text{cls}}=1.0$.

To facilitate better convergence, deep supervision is used so that the loss function is applied at each layer in the transformer decoders. 

\section{EXPERIMENTS}
\subsection{Dataset}
Our approach was tested against the PhenoBench dataset \cite{weyler2023dataset}. Examples of the dataset images can be seen in Figure \ref{fig:examples}. It consists of RGB images of sugarbeet crops and weeds taken from a UAV. These images were annotated on three levels: first plants, weeds and soil was semantically segmented, second plant (crop and weed) instances were segmented, and finally each leaf instance of the sugarbeet crops was segmented. The training set contains 1407 images, the validation set contains 772 images and the test set contains 693 images. The images have a resolution of $1024\times 1024$. This work was submitted to the \emph{PhenoBench: Hierarchical Panoptic Segmentation competition}. Further details about the dataset collection and annotation process can be found in \cite{weyler2023dataset}, and details on the competition can be found at \cite{cvppa}.

\subsection{Training Settings}
Due to the limited training data available, we used a model pretrained on COCO from \cite{cheng2021mask2former}. The backbone used in our experiments was ResNet-50 \cite{he2015deep}. Once again, due to data constraints, the first 4 stages of the ResNet-50 backbone were frozen so that only layers in the final stage were fine tuned on the data. We use Detectron2 \cite{wu2019detectron2} and, as suggested for Mask2Former \cite{cheng2021mask2former}, we use the AdamW \cite{loshchilov2019decoupled} optimizer and the step learning rate schedule. We use an initial learning rate of 0.0001 and a weight decay of 0.05. A learning rate multiplier of 0.1 is applied to the backbone and we decay the learning rate at 0.9 and 0.95 fractions of the total number of training steps by a factor of 10. The model was trained for 100 epochs.


\begin{figure*}[b]
\centering
\fontsize{8}{10}
\subcaptionbox{RTX 2080\label{fig:speed:2080}}
{\includegraphics[width=0.49\textwidth]{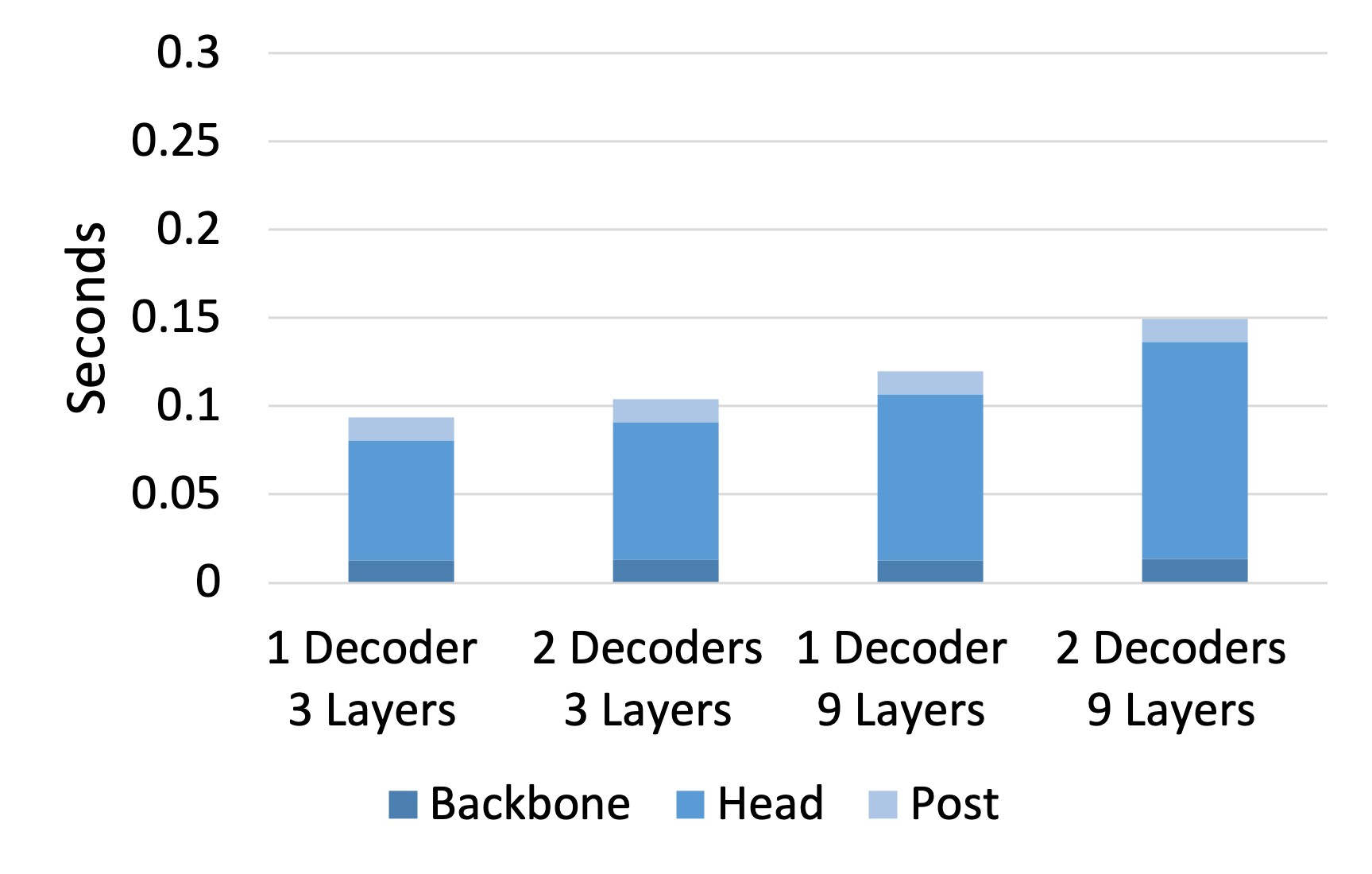}}
\subcaptionbox{Tesla T4\label{fig:speed:t4}}
{\includegraphics[width=0.49\textwidth]{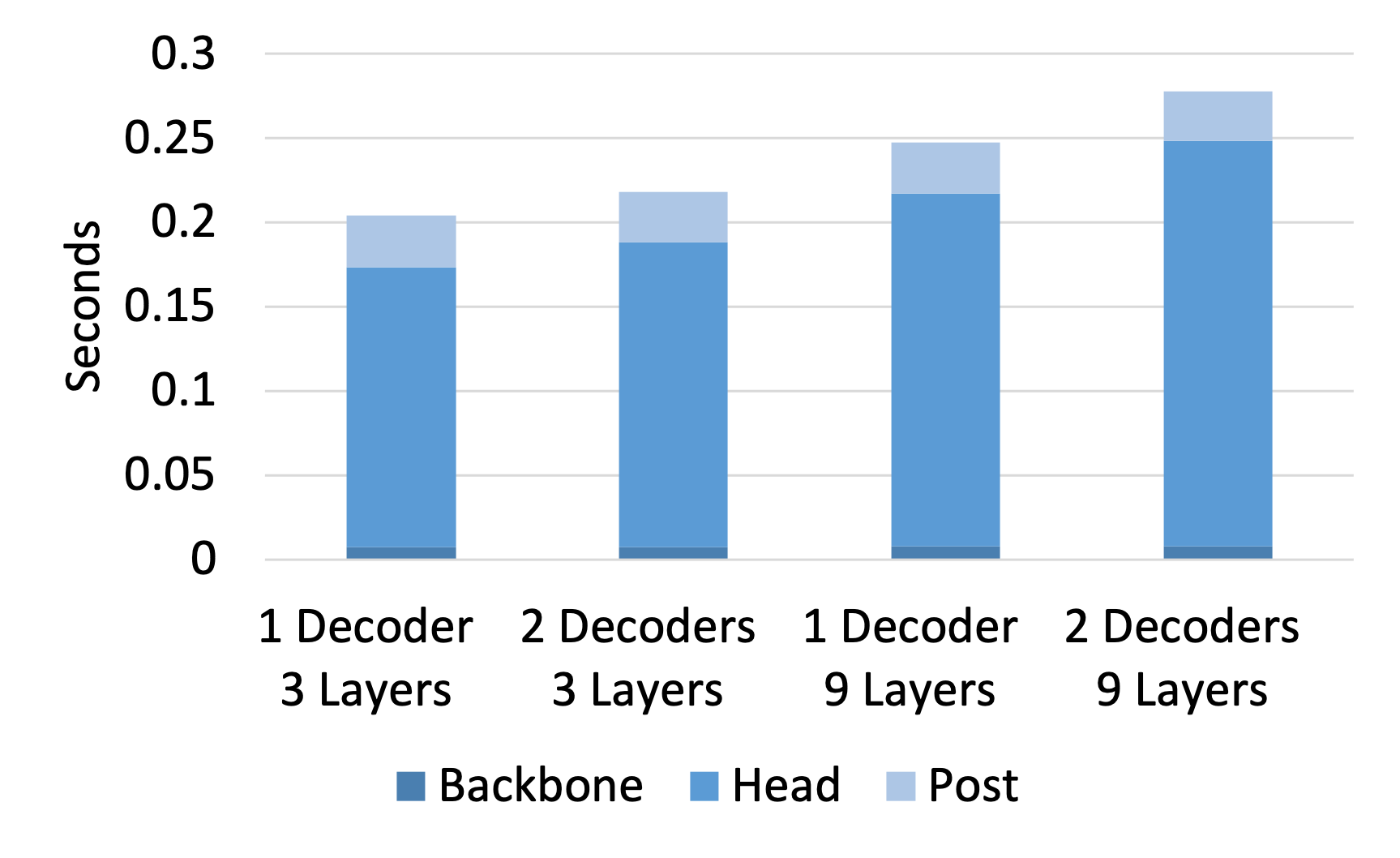}}
\caption{Inference speed results. The time taken to process a single frame, given a batch size of 1, at each stage in the model.}
\label{fig:speed}
\end{figure*}

\subsection{Metrics}
As in \cite{weyler2022field}, panoptic quality \cite{kirillov2019panoptic} is used to assess the predicted masks of crops \(PQ_{crop}\) and leaves \(PQ_{leaf}\) . The average over these values is reported as $PQ$. During evaluation, plant or leaf instances where less than 50\% of it's pixels are within the image, do not affect the score, since these are regarded as uninformative. Additionally, the IoU is calculated for the ``stuff'' categories: weeds \(IoU_{weed}\) and soil \(IoU_{soil}\). The metric \(PQ^\dagger\) is the average over \(PQ_{crop}\), \(PQ_{leaf}\), \(IoU_{weed}\) and \(IoU_{soil}\). 

\subsection{Speed Test}
The models were tested on two GPUs: an Nvidia GeForce RTX 2080 Ti and an Nvidia Tesla T4. The Nvidia Tesla T4 was chosen because it is the type of hardware that could run on a large agricultural vehicle, such as a tractor, in the field. It is available in a ruggedized case with an integrated battery which makes it suitable for field deployment. The inference speed was calculated as the mean time to process each frame, given a batch size of 1. The average inference speed was calculated for each stage of the model: the backbone, the head (which includes the pixel decoder, transformer decoder and segmentation modules), and post-processing. Additionally, the sum of the inference speeds over stages was converted to frames per second (FPS).

\subsection{Inference Settings}
During inference, the binary masks predicted by Mask2Former are filtered to remove those with low confidence predictions and those that are heavily occluded by other masks. This filtering is determined via a mask confidence threshold and an overlap threshold and in our experiments they were set to 0.5 and 0.8, respectively.

\section{RESULTS}
 The inference speed for each stage in the model, on each GPU, is shown in Figure \ref{fig:speed}. The overall inference speed in frames per second, along with the size of each model, is in Table \ref{tab:speed}. A set of example outputs compared to the ground truth is shown in Figure \ref{fig:output}. Results for all four models against the hidden test set compared with other benchmarks are presented in Table \ref{tab:test}.

\subsection{Inference Speed}
Figure \ref{fig:speed} shows that the segmentation head (i.e, the pixel decoder, the transformer decoder and segmentation module) is by far the most time-consuming stage of the model. Reducing the number of layers in the transformer was shown to reduce the inference speed of this stage by a modest amount. Furthermore, adding an additional transformer decoder only results in a small decrease in inference speed.

The frames per second for the smallest model is 4.9 on the Tesla T4 while it's 10.70 on the RTX 2080. Compared the the largest model, this is a 37\% and a 60\% speed increase for the Tesla T4 and RTX 2080, respectively.

\begin{table}[!htb]
\centering
\renewcommand{\arraystretch}{1.1}
\scriptsize
\caption{Model Size and Inference Speed}
\label{tab:speed}
\begin{tabular}{|c|c|c|c|c|}
\hline
    \begin{tabular}{@{}c@{}}Number of \\ Transformers\end{tabular}  & \begin{tabular}{@{}c@{}}Transformer \\ Decoder Depth\end{tabular} & Params & \begin{tabular}{@{}c@{}}FPS \\ (RTX 2080)\end{tabular}  & \begin{tabular}{@{}c@{}}FPS \\ (Tesla T4)\end{tabular} \\ \hline

    1 & 3 & 34.7M & 10.70 & 4.90 \\ \hline
    1 & 9 & 44.1M & 8.37 & 4.04 \\ \hline
    2 & 3 & 39.4M & 9.63 & 4.58 \\ \hline
    2 & 9 & 58.4M & 6.70 & 3.60 \\ \hline

\end{tabular}%
\end{table}

\subsection{Qualitative Analysis}
\begin{figure*}[t]
\centering
\vspace{0.5cm}
\begin{subfigure}{0.99\linewidth}
    \centering
    \includegraphics[width=0.24\linewidth]{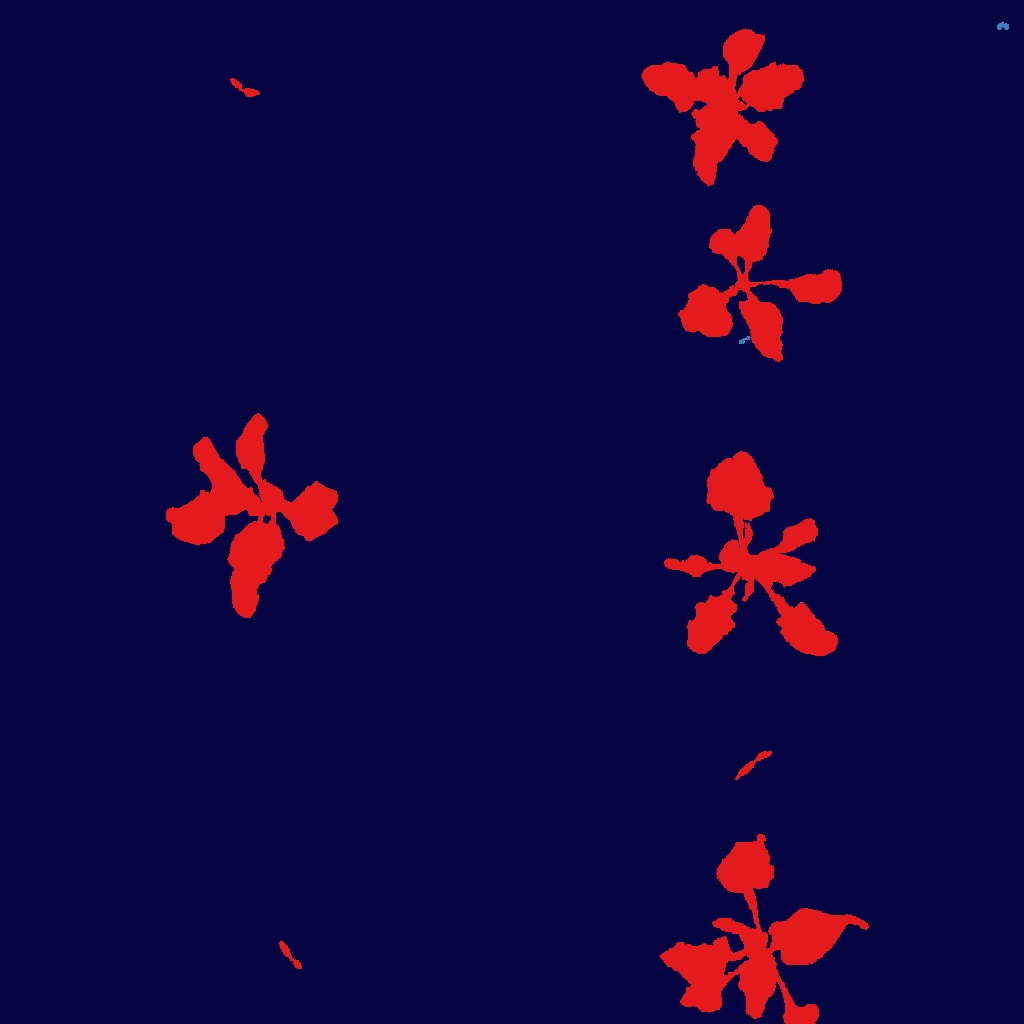}
    \includegraphics[width=0.24\linewidth]{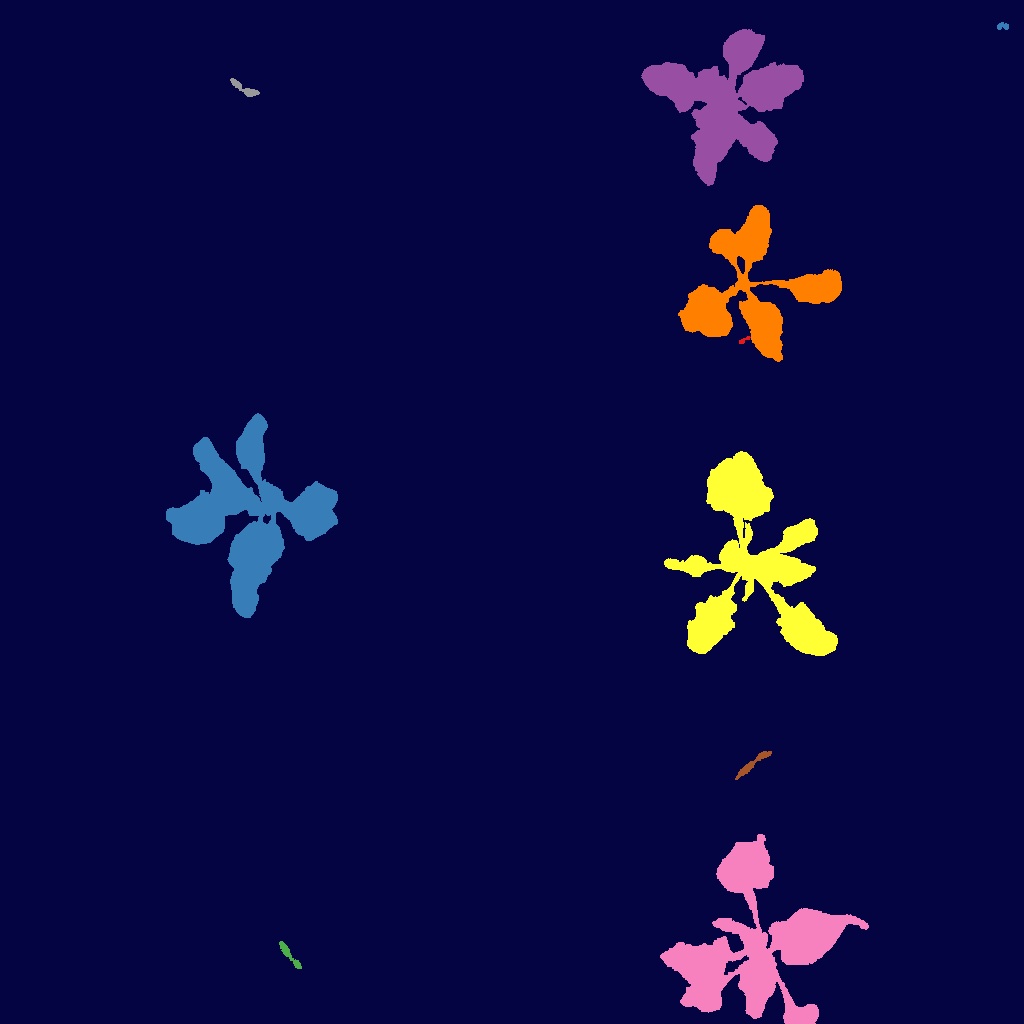}
    \includegraphics[width=0.24\linewidth]{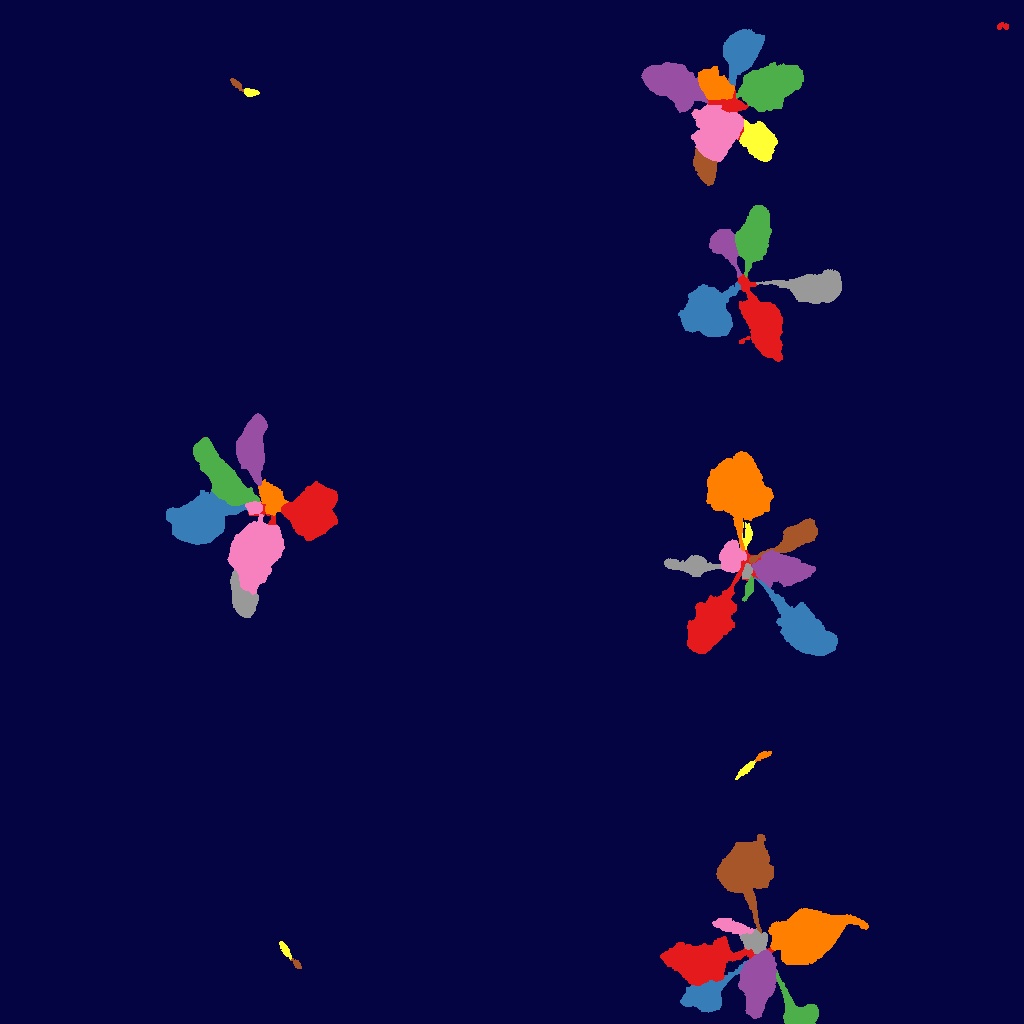}
    \includegraphics[width=0.24\linewidth]{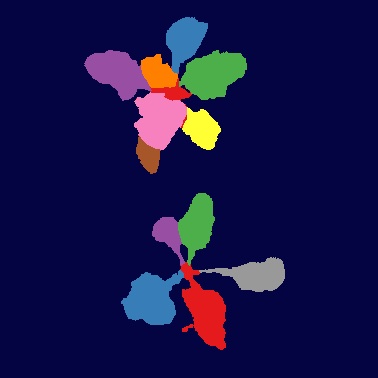}
    \caption{Ground Truth}
\end{subfigure}\vspace{0.3cm}
\begin{subfigure}{0.99\linewidth}
    \centering
    \includegraphics[width=0.24\linewidth]{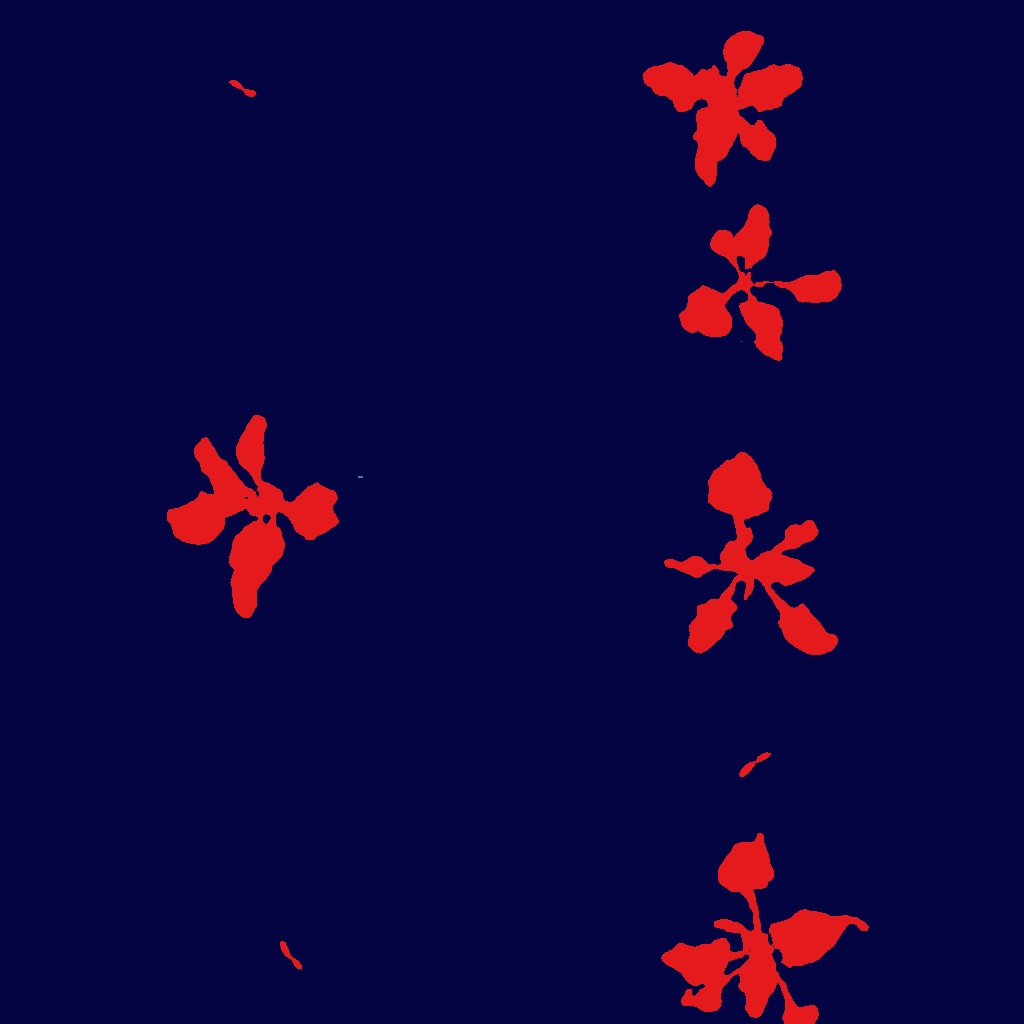}
    \includegraphics[width=0.24\linewidth]{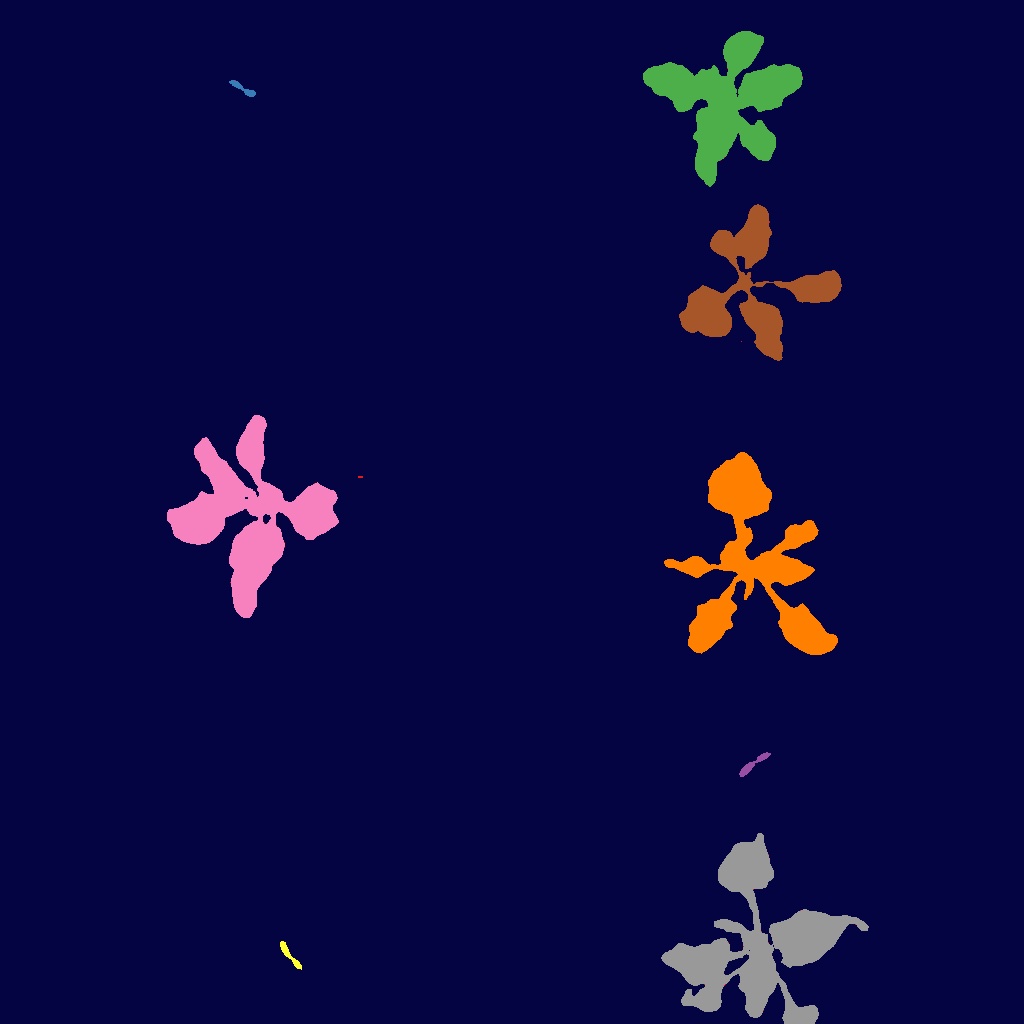}
    \includegraphics[width=0.24\linewidth]{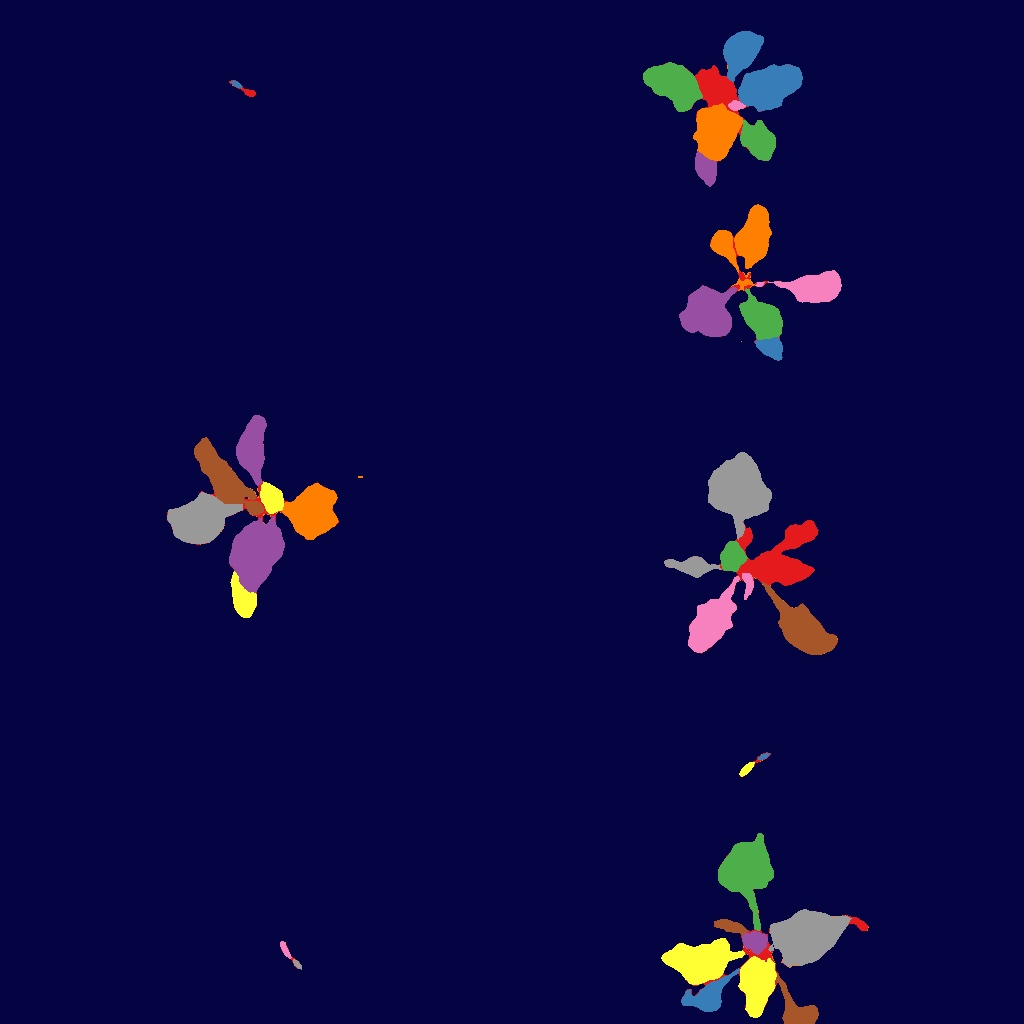}
    \includegraphics[width=0.24\linewidth]{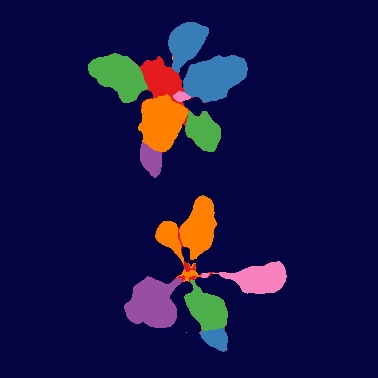}
    \caption{Predictions}
\end{subfigure}\vspace{0.3cm}
\caption{Example outputs from the model with 2 transformers and 9 layers compared to the ground truth. From left to right: semantic segmentation, plant instance segmentation, leaf instance segmentation, close-up of leaf instance segmentation.}
\label{fig:output}
\end{figure*}

Figure \ref{fig:output} shows an example of the model's output during inference. The semantic segmentation of the crops is very accurate with the outlines matching well. However, the semantic segmentation ground truth contain two very small weeds (in light blue) which were not detected by the model. Apart from these weeds that weren't detected, the plant instances seem to be well matched and distinguished from one another. Most of the leaf instances match fairly well though there are several where small parts are mis-classified as another instance or a new instance. By way of example, in the close-up image one leaf in the ground truth has been predicted to be two separate leaves.

\subsection{Comparison with Other Benchmarks}
Mask2Former combines state-of-the-art semantic segmentation and instance segmentation, and since our results show that our model was strong in both semantic segmentation and instance segmentation, these demonstrate the benefit of this approach. All of the models that we tested outperformed all the benchmarks in \cite{weyler2023dataset} across all metrics, apart from \(PQ_{crop}\), where slightly better performance was achieved on a standalone plant segmentation Mask2Former model. This indicates that in optimising for both plant and leaf segmentation tasks simultaneously it is possible to degrade the overall performance in a single task.

\(IoU_{soil}\) is close to perfect, being over 99 for all models, however, the \(IoU_{weed}\) is much lower, in the region of 65, for all models. Weeds present more intra-class variation than crops because the class represents a range of species as well as often being smaller than crops. This makes segmenting weeds more challenging. All model achieved a \(PQ_{crop}\) of over 70 but leaf instances proved more challenging with most models only achieving a \(PQ_{leaf}\) of around 65. However, the \(PQ_{leaf}\) was significantly more accurate than other baselines.

The architectures with more layers tend to better segment weeds while the architectures with two transformer decoders seem to better segment leaves.

\begin{table}[!htb]
\centering
\renewcommand{\arraystretch}{1.1}
\scriptsize
\caption{Comparison of our results on the hidden test set with others published in \cite{weyler2023dataset}. Our models have the highest performance (in bold) on all but one metric. Note the abbreviation in our models e.g., (1T, 3L) refers to a model with 1 Transformer with 3 Layers.}
\label{tab:test}
\begin{tabular}{|c|c|c|c|c|c|c|c|c|}
\hline
  Model & PQ+ & PQ & \begin{tabular}[c]{@{}c@{}} \(PQ_{crop} \)\end{tabular}  & \begin{tabular}[c]{@{}c@{}} \(PQ_{leaf} \)\end{tabular} & \begin{tabular}[c]{@{}c@{}} \(IoU_{weed} \)\end{tabular} & \begin{tabular}[c]{@{}c@{}} \(IoU_{soil} \)\end{tabular} \\ \hline
\begin{tabular}[c]{@{}c@{}} ERF-Net\end{tabular} & - & - & - & - & 64.37 & 99.28 \\ \hline
\begin{tabular}[c]{@{}c@{}} DeepLabV3\end{tabular} & - & - & - & - & 64.59 & 99.25 \\ \hline
\begin{tabular}[c]{@{}c@{}} Mask R-CNN\end{tabular} & 65.79 & - & 67.61 & - & - & 98.47 \\ \hline
\begin{tabular}[c]{@{}c@{}} Mask2Former\end{tabular} & 69.99 & - &  \textbf{71.21} & - & - & 98.38 \\ \hline
\begin{tabular}[c]{@{}c@{}} Mask R-CNN\end{tabular} & - & - & - & 59.74 & - & - \\ \hline
\begin{tabular}[c]{@{}c@{}} Mask2Former\end{tabular} & - & - & - & 57.50 & - & - \\ \hline
\begin{tabular}[c]{@{}c@{}} Weyler \cite{weyler2022field} \end{tabular} & - & 40.43 & 38.37 & 42.60 & - & -  \\ \hline
\begin{tabular}[c]{@{}c@{}}HAPT\cite{roggiolani2023hierarchical} \end{tabular} & 65.27 & 50.73 & 54.61 & 46.84 & 61.11 & 98.50  \\ \hline
\begin{tabular}[c]{@{}c@{}}\textbf{Ours} (1T, 3L)\end{tabular} & 75.01 & 67.38 & 70.16 & 64.61 & 65.94 & 99.34 \\ \hline
\begin{tabular}[c]{@{}c@{}}\textbf{Ours} (1T, 9L)\end{tabular} & 75.25 & 67.49 & 70.61 & 64.37 & \textbf{66.67} & 99.33 \\ \hline
\begin{tabular}[c]{@{}c@{}}\textbf{Ours} (2T, 3L)\end{tabular} & 75.1 & 67.8 & 70.18 & 65.41 & 65.41 & 99.35 \\ \hline
\begin{tabular}[c]{@{}c@{}}\textbf{Ours} (2T, 9L)\end{tabular} & \textbf{75.99} & \textbf{69} & 71.1 & \textbf{66.91} & 66.58 & \textbf{99.36} \\ \hline
\end{tabular}%
\end{table}

\section{DISCUSSION}

\subsection{Architecture}
There appears to be a trade-off between the different architectures used. The double transformer decoder models achieving a higher \(PQ_{crop}\) and the single transformer decoder models achieving a higher \(IoU_{weed}\). It's possible the leaf segmentation benefits from the transformer decoder becoming specialised at that level of the segmentation. By contrast, it seems optimising the second decoder means the network does not optimise as well for the weed class. Further work could investigate how to adapt the loss function to better train the double transformer model without degrading the weed class accuracy.

\subsection{Inference Speed}
There was an improvement in inference speed from making the network more compact. The smallest architecture ran at almost 5 FPS on the Tesla T4 and over 10FPS on the RTX 2020. This will not be fast enough for many applications, but considering the high resolution of the images used, this represents competitive performance. One solution is to reduce the resolution but this would likely degrade accuracy. Our results show that the inference speed overhead is mostly within the segmentation head, and that even reducing layers in the transformer decoder did reduce this dramatically. Future work will focus on how to make more efficiencies in the segmentation head to speed up inference.

\section{CONCLUSIONS}
Overall, this work demonstrates the potential of architectures like Mask2Former to solve the visual recognition challenges within precision agriculture. We demonstrate that it can be easily adapted for hierarchical panoptic segmentation. Additionally, our work shows that a compact implementation can still achieve high accuracy. In future work, we would like to investigate how to further reduce the size of the proposed model while retaining its accuracy.

\bibliographystyle{IEEEtran}
\bibliography{IEEEabrv,references}

\end{document}